\documentclass[lettersize,journal]{IEEEtran}
\usepackage{multirow}
\usepackage[table,xcdraw]{xcolor}
\usepackage{amsmath,amsfonts}
\usepackage{algorithmic}
\usepackage{algorithm}
\usepackage{array}
\usepackage[caption=false,font=normalsize,labelfont=sf,textfont=sf]{subfig}
\usepackage{textcomp}
\usepackage{stfloats}
\usepackage{url}
\usepackage{verbatim}
\usepackage{graphicx}
\usepackage{cite}
\usepackage{xcolor}  % Optional: for syntax coloring
\usepackage{changepage}
\usepackage{hyperref}
\usepackage[most]{tcolorbox}
\usepackage{lipsum}

\newtcolorbox[auto counter, number within=section]{convbox-prompt}[1][]{colframe=blue!50!white,colback=blue!10!white,
  coltitle=black, fonttitle=\bfseries, title= User,boxrule=0.8mm, width=\columnwidth, sharp corners, #1, breakable}

\newtcolorbox[auto counter, number within=section]
{convbox-response}[1][]{colframe=black!50!white,colback=black!10!white,
  coltitle=black, fonttitle=\bfseries, title= Assistant,boxrule=0.8mm, width=\columnwidth, sharp corners, #1, breakable}

\newtcolorbox[auto counter, number within=section]
{convbox-response-inj}[1][]{colframe=black!50!white,colback=black!10!white,
  coltitle=black, fonttitle=\bfseries, title= Assistant (Injected) ,boxrule=0.8mm, width=\columnwidth, sharp corners, #1, breakable}

\begin{document}

% \title{Language-based Analysis of Synergies in Dynamic Rule Systems}
%\title{Language-based Reasoning for Predicting Synergies in Dynamic Rule Systems}
\title{Rule Synergy Analysis using LLMs: State of the Art and Implications}


\author{Bahar Bateni, Benjamin Pratt, and Jim Whitehead
\thanks{The authors are with the Jack Baskin School of Engineering, University of
California Santa Cruz, Santa Cruz, CA 95064 USA (e-mail: bbateni@ucsc.edu;
bepratt@ucsc.edu; ejw@ucsc.edu).}}

% \author{authors}

% TODO


\maketitle

\begin{abstract}
Large language models (LLMs) have demonstrated strong performance across a variety of domains, including logical reasoning, mathematics, and more. In this paper, we investigate how well LLMs understand and reason about complex rule interactions in dynamic environments, such as card games. We introduce a dataset of card synergies from the game \textit{Slay the Spire}, where pairs of cards are classified based on their positive, negative, or neutral interactions. Our evaluation shows that while LLMs excel at identifying non-synergistic pairs, they struggle with detecting positive and, particularly, negative synergies. We categorize common error types, including issues with timing, defining game states, and following game rules. Our findings suggest directions for future research to improve model performance in predicting the effect of rules and their interactions.
% Games with dynamic rule systems allow the players to enable or disable certain rules during gameplay. In such systems, the rules are designed in a way to enable a vast number of synergistic effects arising from combining different rules.

% In this paper, we propose the use of LLMs for evaluating rules in terms of potential for synergies. We introduce a dataset of cards and card synergies for the roguelike card game \textit{Slay the Spire}, and demonstrate the use of LLMs for predicting these synergy effects without playing the game. Furthermore, we classify the common errors in these predictions, and provide a comparison of different prompt types and models in terms of success in predicting synergies.
\end{abstract}


% \textit{TODO: Static Analysis vs Dynamic Analysis. The importance of static analysis which is novel about this work. Also give a broader relevance argument: we want really robust automatic systems about creating games. They need the ability to reflect on impact of rules and their interactions. This is not just a classification task using LLMs. This is broader and is aimed for systems that design rules and reflect on them.}

\section{Introduction}

Large language models (LLMs) have shown promising results in performing a wide range of language and reasoning tasks. Recent benchmarks have demonstrated their abilities in logical reasoning, mathematics, coding, and more. These models have also been used in games, where they have proven capable of tackling various tasks, such as level generation, game-playing, NPC role-playing, automated game design, and even acting as a game master \cite{llms-in-games}. Notably, these tasks sometimes require a deep understanding of game design principles.

Given the demonstrated proficiency of LLMs in these areas, we were curious to investigate their ability to understand and reason about game rules, particularly focusing on how different game rules interact and produce emergent effects. This is an important area of exploration, as many games—especially those with dynamic rules, such as collectible card games—rely on rule interactions to provide depth and strategic variety. To evaluate this, we designed a task aimed at assessing how well LLMs can reason about synergies between different game rules, specifically in the context of card games.

Card games are a rich domain for studying rule interactions, as they often feature a combination of simple, discrete rules that create complex interactions when combined. Each card can describe various effects, and when played together, cards may produce synergy effects that either positively or negatively interact, adding strategic depth to the game. One such game that leverages these synergies extensively is \textit{Slay the Spire} \cite{slay-the-spire}, a roguelike deck-building card game in which players build their decks through random card additions at the end of each battle. As the game progresses, the player’s deck evolves, with each card selection potentially enhancing or undermining synergies with previously acquired cards.

To assess the ability of LLMs to reason about rule synergies, we propose a dataset of card synergies derived from \textit{Slay the Spire}. The dataset includes all pairs of cards from one of the game's card sets, where each pair is classified based on how they synergize: positive, negative, or no synergy. We prompt the LLM to determine the synergy for each pair, and we provide the ground truth values for this dataset, which we have made publicly available\footnote{The dataset, fine-tuning examples, and the source code can be found at \href{https://github.com/iambb5445/Language-based-Synergy-Detection}{\url{https://github.com/iambb5445/Language-based-Synergy-Detection}}}. This task is particularly interesting as it requires an understanding of both the individual card effects and the complex interactions that emerge when cards are combined. Understanding these synergies not only requires interpreting the specific effects of each card but also demands reasoning about the different types of game logic, which we explore further in Section~\ref{sec:challenges}.

In addition to testing LLMs' understanding of game rules, this task has significant practical implications for the field of game design. Analyzing synergies in this manner could contribute to the development of mixed-initiative design tools or robust automated game design systems capable of evaluating rules statically. This means that without having a playable implementation, the system can reflect on the impact of adding a new rule and its possible interactions with existing rules. In board game design, such reasoning could permit a design system to provide feedback on rule systems and make design suggestions. In collectible card game design, such systems could assess the strategic implications of new rules, for example predicting how a newly generated card would interact with an existing set. It could also help designers optimize card sets to encourage specific types of strategic play, such as designing cards that have synergies with all other cards (for utility) or cards that are more situational and can cause both positive and negative synergies.

Furthermore, our analysis provides insights into the types of mistakes LLMs make when reasoning about rule synergies. Although LLMs tend to perform well in determining when there is no synergy, they struggle to identify positive and negative synergies. We classify several common error types, such as misunderstandings related to the timing of card effects, following rules of the game, and envisioning unrealistic game states. These insights highlight areas where current LLMs still fall short and suggest directions for future research to improve the models' ability to reason about complex game rules. Our findings contribute to the broader conversation on the use of LLMs to assist in the development of interactive systems that require a deep understanding of complex rule-based systems.

\section{Related Work}

LLMs have found diverse applications in the context of games. In performing Procedural Content Generation (PCG) tasks, LLMs have shown promise in generating game components that require an understanding of game design principles. For instance, MarioGPT \cite{mariogpt} utilizes a fine-tuned version of GPT-2 to generate levels for Super Mario Bros (SMB). The model produces a level as a text string based on a simple input prompt that specifies the frequency of various game elements, and also verifies the playability of the generated levels by generating a path for the player. To ensure diversity in the generated levels, MarioGPT employs a novelty search genetic algorithm.

Another notable example of LLM-based level generation is the work by Todd et al. \cite{llm-sokoban}, which focuses on generating levels for the puzzle game Sokoban. Similar to MarioGPT, the levels are represented as text, and both GPT-2 and GPT-3 are used to create diverse and playable levels. The authors highlight that while LLMs excel at generating levels based on simple instructions, such as the frequency of empty tiles, they struggle with more complex design metrics like the solution length. Our own findings corroborate this, as we see that LLMs perform poorly when tasked with reasoning about intricate aspects of game design, such as predicting complex rule interactions.

In the domain of card games, LLMs have been employed in various ways, with some systems focusing on generating thematically relevant cards, but without direct consideration for game design balance. For example, Cardistry \cite{cardistry} uses GPT-3.5 to convert a short personal story into a playing card. The model generates card attributes and prompts for image generation, ensuring that the card thematically represents the story. Similarly, CrawLLM \cite{craw-llm} adapts a card game to a desired setting and narrative by using Mixtral to retheme the cards. Both works use the creative potential of LLMs to generate cards based on thematic storytelling, rather than addressing the design of gameplay rules. In contrast, our focus is on reasoning on evaluation of design elements.

Language-based methods have also been used to assist in game design, particularly in creating cards with interesting rule effects. One of the earliest examples is RoboRosewater \cite{roborosewater}, an automated card generation system based on a deep Recurrent Neural Network (RNN). RoboRosewater's primary goal is to imitate the text representations of cards in \textit{Hearthstone}, without considering game balance or the interactions between cards. This results in the generation of creative and novel cards, but many of them are unbalanced or even unplayable.

Mystical Tutor \cite{mystical-tutor} is a mixed-initiative tool that aids in generating cards for \textit{Magic: The Gathering} by completing partial card descriptions. Mystical Tutor models this task as a translation problem to fill-in the missing information in a partial description. Similar to RoboRosewater, while this can result in creative and interesting cards, Mystical Tutor does not automatically evaluate the generated cards for playability, relying instead on human judgment and accuracy metrics of the translation process.

Chaos Cards \cite{chaos-cards} uses a grammatical model to generate both the card descriptions and a playable version of the card, which allows the system to evaluate the cards through a simulation-based approach. In our work, we focus on a static evaluation approach, where the evaluation does not rely on the existence of a playable version of the card.

While the idea of static evaluation has potential benefits, our results show that using LLM agents for this type reasoning may result in poor performance. While LLMs have shown more than acceptable results in many other tasks, game reasoning related tasks still remain an area of improvement. For example, BoardgameQA \cite{boardgame-qa}, a benchmark dataset for evaluating LLMs in the context of board games, demonstrates that LLMs struggle when tasked with answering questions about a game given its state and rules. It should be noted that one of the main focuses of the BoardgameQA dataset is providing incomplete or contradictory data and evaluating LLMs on their ability to recover from these cases. However, one challenge of our work is that card synergy detection and, more generally, reasoning about game design rules requires reasoning about general implications of a rule, which means that the LLM must infer the broader implications of game rules without explicit game state information.

Another area where LLMs fall short is in temporal reasoning. Research \cite{temporal-llm} has shown that LLMs perform poorly on tasks requiring the understanding of event sequences or temporal dependencies. Similarly, we have found that errors in reasoning about the order of events are one of the most common types of mistakes made by LLMs in the context of game design.

Finally, our previous work \cite{llm-as-ggp} demonstrated that LLMs can play the game \textit{Slay the Spire} with acceptable performance, particularly when long-term planning is required. However, our task in this paper differs in that there is no game state to guide the LLM. This means that the model must generate coherent gameplay decisions without any information about game state, which poses a unique challenge. Another reason is that when a game state is simulated in the process of game-playing, it provides immediate feedback after every move on the correct state of the game, clearing timing and rule errors.

% \section{Dynamic Rule Systems}

% \textit{TODO: possibly combine with introduction.}

% \textit{TODO: This section will describe dynamic rule systems, what they are, and why they are important. I don't want to completely repeat what I already had in the MiniStS paper, so I'll include only a summary here. I'll also include a revised version of one of the figures from that paper (Fig.~\ref{fig:dynamic-rule-system} - revised based on Ben's feedback. This section will also briefly introduce Slay the Spire.}

% \begin{figure}
%     \centering
%     \includegraphics[width=\linewidth]{dynamic-rule-system.pdf}
%     \caption{Visualization of a dynamic rule system. The initial set of composable rules are created by the designer. During a playthrough, the player impacts which rules are in play.}
%     \label{fig:dynamic-rule-system}
% \end{figure}

\section{Synergy Definition}

A synergy effect is defined as the interaction of two or more elements to produce a combined effect greater than the sum of their individual effects. In the context of card games, this can manifest as cards interacting with each other to produce a more powerful or advantageous outcome than when played independently. For instance, consider two cards: Card A, which grants “Gain 5 Block,” and Card B, which “deals damage equal to your Block.” While each card has its standalone value, when used together the outcome is enhanced. Playing Card A first provides the player with Block, and then playing Card B leverages that Block to deal additional damage. This is an example of positive synergy, where the combined effect is greater than the sum of the effects of the individual cards, creating an advantage.

In contrast, a negative synergy is when two cards interact in ways that diminish or counteract each other’s intended effects. For example, Card A could be “Prevent the next time you receive damage,” while Card B might be “Costs 1 less energy for each time you lose HP in combat. Deal 18 damage.” While card A is normally considered positive since it’s preventing incoming damage, when played with card B it has an added negative implication which is preventing the cost reduction of card B. This creates a negative synergy, where the effects of the cards are less beneficial together.

Synergy effects play an important role in adding strategic depth to a game. The decision of whether to play a card or include it in a deck becomes significantly more complex when synergy effects are considered. A card's inherent value is no longer static but varies depending on the other cards in the deck and the broader game state. For instance, a card that seems mediocre in isolation might be highly valuable when synergized with others. As such, players must carefully consider not just the individual value of a card, but its potential interactions within the context of the entire deck or the evolving game state.

Moreover, synergy effects contribute to what is known as Lenticular Design, a concept introduced by Mark Rosewater, the lead designer of \textit{Magic: The Gathering}. Lenticular Design refers to the ability of a game to become more complex as a player gains experience. This is desirable because new players can quickly grasp the fundamentals and begin enjoying the game, while experienced players can continue to discover deeper layers of strategy and nuance. Synergies contribute to Lenticular Design, as they introduce interactions that reveal new possibilities as the player becomes more familiar with the game. A novice player might understand a card’s effect in isolation, but an experienced player can anticipate the potential synergies it has with other cards, unlocking new avenues for how to play the game.

Finally, synergy effects can create a wide range of ways the game can be played. In card games, even a limited set of cards can result in many possible decks which play differently. In many roguelikes, where each playthrough is designed to offer a unique experience, the game includes a set of rules that the player can choose from and combine together to create a different play experience every time. Synergies help with this goal because any set of rules may have additional effects that can make the game play differently.

\section{Synergy Analysis Dataset}

Detecting synergies between game elements demands a deep understanding of game design principles. The ability to evaluate how different rules and mechanics interact with each other is important for both designing and evaluating games. By assessing a system’s ability to detect synergies, we can understand its capacity for predicting complex interactions within a game. If a system can accurately detect synergies, it could be used for several purposes, such as evaluating rule sets in a rule generation task or serving as a mixed-initiative tool for game design. For instance, a game designer could leverage such a system to explore potential card combinations, helping to optimize a game's balance.

In this study, we use the synergy analysis dataset to evaluate large language models (LLMs) on their ability to detect synergies statically. Static evaluation requires only the descriptions of the cards and does not rely on a playable version of the cards. In contrast, dynamic evaluation—where the system can experiment with the cards in a testbed—relies on running the game to observe interactions. Static evaluation is a challenging task because the system must rely purely on reasoning about the textual descriptions of card effects, without access to the immediate feedback from gameplay. We discuss the specific challenges of this task further in Section~\ref{sec:challenges}.

To facilitate this evaluation, we introduce a Synergy Analysis Dataset based on one of the card sets from the game \textit{Slay the Spire}. This card set consists of 75 cards, and the dataset includes the synergy value for each pairwise combination, resulting in a total of 5,625 pairs. For each pair, the dataset labels the interaction as one of three categories: positive synergy (denoted as +1), negative synergy (denoted as -1), or no synergy (denoted as 0). Table~\ref{tab:dataset-classes} shows the distribution of synergies across the dataset, detailing the number of positive, negative, and neutral (no synergy) pairs. As you can see, the dataset include significantly smaller number of negative synergies, since this type of synergy is less common. Because of this, while we include the accuracy measure on all tasks, our main measurement for evaluating the result is the F1 score.

\begin{table}[!t]
\caption{Frequency of each class in the synergy analysis dataset\label{tab:dataset-classes}}
\centering
\renewcommand{\arraystretch}{1.4}
\begin{tabular}{|c||c|c|}
\hline
Class & Count & Percentage\\
\hline
Positive Synergy & 947 & 16.8\%\\
No Synergy & 4553 & 81\%\\
Negative Synergy & 125 & 2.2\%\\
\hline
Total & 5625 &  \multicolumn{1}{c}{}\\
\cline{1-2}
\end{tabular}
\end{table}

A synergy between card A and card B is defined as whether playing card A first and card B after results in an effect that differs from the sum of their individual effects. Specifically, we detect whether the combined effect of playing card A first and then Card B provides an advantage (positive synergy) or a disadvantage (negative synergy) or no change (no synergy) to the player. It is important to note that some synergies may be delayed or long-term in nature. For example, a card might have an effect that does not immediately trigger, but can enhance another card's effect later in the game. Therefore, the dataset accounts for synergies that unfold over multiple turns, rather than requiring the effects to be immediate.

\begin{figure*}
    \centering
    \includegraphics[width=0.9\linewidth]{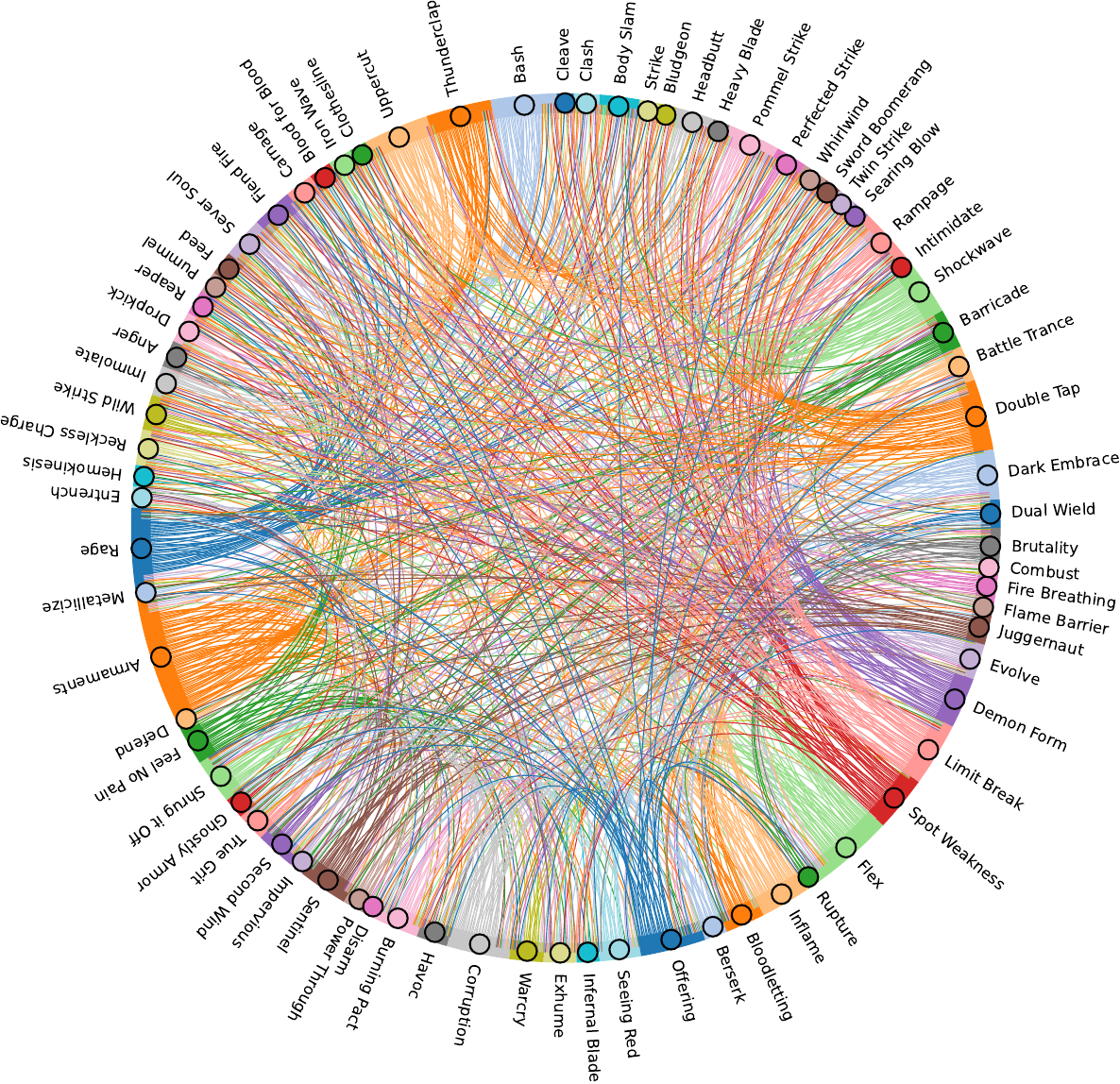}
    \caption{Visualization of all the positive synergies in the dataset. Each dot represents a card, and the cards are sorted by their effect types. For example, cards that add strength (add a flat damage to attacks) are located at the bottom, starting from ``Demon Form'' to ``Inflame''. These cards synergizes with all the cards that deal damage, which are located at the top starting from ``Rampage'' to ``Reckless Charge''.}
    \label{fig:chord-diagram}
\end{figure*}

Figure~\ref{fig:chord-diagram} visualizes the positive synergies within the game, highlighting clusters of cards that exhibit synergy patterns with each other. These patterns can often be observed in groups of cards that share mechanical similarities, such as cards that focus on blocking, applying damage, or damage amplification. These groups of cards tend to have similar synergy values with other cards their respective categories, suggesting that their effects complement one another in. to some extent, predictable ways.

\section{Challenges}
\label{sec:challenges}

We previously defined the synergy detection task which requires statically analyzing the cards, meaning that our aim is to evaluate systems on their ability to predict the way card effects interact with each other without using a testbed to simulate these interactions. This makes it a particularly difficult task, even for humans, as it involves not only interpreting game rules and card effects, but also understanding how they interact in complex ways. We now discuss the main challenges associated with this task.

\subsection{Rule interpretation}
\label{sec:challenges-rule-interpretation}

Each card is described using natural language, aiming to clearly communicate its effects. However, while the descriptions are carefully crafted to avoid any confusion, it is possible for the descriptions to be misunderstood.
Misunderstandings can arise due to ambiguities in the card descriptions or differences in player experience. For example, the card “Sword Boomerang” is defined as “Deal 3 damage to a random enemy 3 times.” This description does not clearly indicate whether this random enemy is the same for all 3 times the damage is inflicted, or if each time an enemy is chosen by random and the damage can be applied to a different enemy each time. While the card description may seem clear, subtle differences in wording or phrasing can lead to confusion.

Additionally, players’ prior experience with other games can also influence their interpretation of card effects. In \textit{Slay the Spire}, the effect “shuffle the card into your draw pile” means that the card is placed in a random position within the draw pile. However, in other card games like \textit{Hearthstone}, “shuffle into” may imply shuffling the entire deck. Such cross-game misinterpretations can cause errors in synergy detection, even when the card descriptions themselves are technically correct.

\subsection{Mixed Effects}

Many cards feature both positive and negative effects. These cards add complexity to the synergy analysis, as the player must consider how the beneficial and detrimental aspects of the card interact with other cards in the deck. For example, the card “Battle Trance” is defined as “Draw 3 cards. You cannot draw additional cards this turn.” While this effect is generally positive, it also limits the player’s ability to draw more cards during the same turn. These mixed effects in cards make the synergy analysis more complex, since each of these contradicting effects should be considered during the analysis.

\subsection{Conflicting Synergies}

In our dataset, synergy is classified as either positive, negative, or neutral, condensing the effects of a card pair into a single value. However, some card combinations present both positive and negative synergies simultaneously. For instance, consider the card “Wild Strike,” which reads “Deal 11 damage. Shuffle a Wound into your draw pile.” A “Wound” card is unplayable and detracts from the deck’s utility. When paired with another card such as “Battle Trance” defined as “Draw 3 cards. You cannot draw additional cards this turn,” the synergy can be interpreted in two ways:

\begin{enumerate}
    \item Negative synergy: The “Wound” card created by “Wild Strike” reduces the value of “Battle Trance,” as it adds an unwanted, unplayable card to the deck, making it harder to draw useful cards.
    \item Positive synergy: The “Wound” card is less harmful since “Battle Trance” helps the player cycle through cards more quickly, reducing the negative impact of the Wound.
\end{enumerate}

This combination of conflicting effects presents a significant challenge for both human players and models trying to predict synergies accurately.

\subsection{Randomness}

Randomness is an inherent part of card games, where effects may involve probabilistic elements. For example, cards in \textit{Slay the Spire} can “Add a random attack card to your hand”. This introduces uncertainty, as the outcome depends on which specific card is added. Even when a card does not have an inherently random effect, the shuffling of the deck introduces variability in card order, which can alter the outcome of any card effect.

While experienced players can make educated guesses about the potential results of random effects based on knowledge of the game, predicting synergies becomes difficult due to the inherent unpredictability. In some cases, the model must account for expected values and probabilistic outcomes, making synergy detection more complex.

\subsection{Design Subversion}

Card games often subvert players’ expectations of what constitutes a positive or negative effect. For example, while taking damage is typically seen as a disadvantage, some cards in \textit{Slay the Spire} become stronger when the player intentionally takes damage. This subversion of expectations can be confusing, as it requires the player to rethink what would normally be considered negative in a typical game context.

These design choices introduce counterintuitive synergies, where negative effects on one card may synergize positively with other cards, creating complex interactions that are hard to analyze.

\subsection{Resource Management}

Card games often feature intricate resource management systems. This may include managing resources such as energy, health, card draws, and other game variables. Some cards require sacrificing one resource to gain another, introducing a layer of complexity in synergy analysis. For example, a card may sacrifice energy to allow the player to draw more cards, or it might cause the player to lose health in exchange for gaining energy or other benefits.

The interplay between these different resources further complicates the synergy prediction task, as the value of a card’s effect depends not only on the cards around it but also on the broader resource context in the game.

\subsection{Scaling}

The scale of a card's effect can play a role in determining how synergistic it is with other cards. For example, a small increase in damage might have a substantial impact on a low-cost, low-damage card, but it may have little effect on a high-cost, high-damage card.

Another example is how a small Block card might synergize well with another card that benefits from receiving damage, as the Block will prevent some damage but still allow the synergistic effect to trigger. However, a large Block card might completely prevent the damage needed for the synergy to work, creating a negative interaction. These scaling effects depend not only on the cards themselves but also on the specific game state, making synergy detection more challenging.

\subsection{External Factors}
\label{sec:challenge-external-factor}

For simplicity, we define our synergy task between only two cards. However, sometimes synergies can require specific conditions to be met or other cards to be played to become meaningful. For instance, consider the card “Barricade,” which states, “Block is not removed at the start of your turn.” When paired with a card like “Blood for Blood,” which is defined as “Costs 1 less energy for each time you lose HP in combat. Deal 18 damage,” the synergy between the two is not immediately negative. However, after playing “Barricade,” the player is likely to accumulate a substantial amount of Block, which will prevent incoming damage. This in turn reduces the effectiveness of “Blood for Blood,” as it relies on the player losing HP to reduce its cost.

In such cases, analyzing the synergy between the two cards is complicated because the synergy may depend not only on the two cards in question but also on the broader context—specifically, the game state, the player’s resources, and other cards played.

In our definition of synergy, we allow for the possibility of other cards being played between the two or for the player to pass turns to capture long-term synergy effects. However, this introduces ambiguity, as it can be difficult to determine whether the added effect is a result of the interaction between the two cards or due to the broader scenario that makes the synergy possible.

% \textit{TODO: for understanding synergies you have to understand a lot of different game logics that are combined together. It's a hard task because you have to synthesize different things to get to that. Use the term "logic". Card game logics. time, deck-building logic, hand logic, resource management logic, artificial value, ...
% understanding scaling, }

\section{Prompt Structure}

In order to predict the synergistic effect between two cards, we define a structured conversation with the large language model (LLM) for each prompt. This conversation consists of several components, including a context block with essential information about the game and the task, followed by a question and the model's expected response. This section outlines the structure of the prompt and the components expected in the LLM's answer.

\subsection{Context Block}

The context block provides the LLM with background information for understanding the game and task. It consists of the following parts:

\begin{itemize}
    \item \textbf{Game Description:} This section explains key aspects of the game, such as the sequence of events during a turn, general gameplay mechanics (e.g. card drawing, discarding, dealing damage, etc,) and the win/loss conditions. It sets the foundation for understanding how cards interact in the game.
    \item \textbf{Game Keywords:} A list of game-specific terms used in the card descriptions. For example, "Block" in the game is defined as “Prevents damage until the start of your next turn,” and "Vulnerable" means “Takes 50\% more damage from attacks.” These keywords help the LLM interpret the effects described on the cards.
    \item \textbf{Synergy Definition:} A clear definition of synergy between two cards, which is described as a combined effect that results from playing both cards in a specific order. This section clarifies that the synergy could involve immediate card plays or require actions over multiple turns. Additionally, it specifies the possible outcomes for synergy: "1" for positive synergy, "-1" for negative synergy, and "0" for no synergy.
    \item \textbf{Few-shot Examples:} A set of example question-answer pairs demonstrating the task. These examples help guide the LLM by showing the format of both the question and the expected response. We provide 8 examples to establish the expected flow and structure of the conversation.
\end{itemize}

\subsection{Question Structure}
\label{sec:prompt-structure-question}

The question section provides the LLM with the necessary details for evaluating the synergy between two cards. For each card involved, the prompt includes a complete description of the card's type, cost, and effect. Importantly, we omit the card names to reduce potential biases that could arise from prior associations with those names or the game, as shown in our previous work \cite{llm-as-ggp}. Instead, cards are referred to by numbers, ensuring that the LLM focuses mainly on the card effects rather than any preconceived notions about specific cards.

\subsection{Expected Answer Format} 
\label{sec:prompt-structure-answer}

This format is explicitly defined in the prompt and demonstrated with few-shot examples. The expected response includes several steps of reasoning, which we find crucial to avoid errors and ensure accurate synergy analysis.

This section describes the format that is expected in LLM's response, which includes multiple examples to enable chain-of-thought reasoning. This format is explicitly defined in the prompt and demonstrated with few-shot examples. We found it important to specifically title each of these sections and include all of them in the few-shot examples to prevent "Skip Error", which we later discuss in Section~\ref{sec:skip-error}.

\begin{itemize}
    \item \textbf{Card Description:} The first section of the response is a recitation of the descriptions of the cards involved. This step is important because it reinforces the model's understanding of the card effects, reducing the likelihood of "Description Errors" (see Section~\ref{sec:description-error}).
    \item \textbf{Order of Events:} The next section in the response describes the sequence of events when the cards are played in the given order. This includes explaining what happens when the first card is played, followed by the effects of the second card. This step ensures that the model considers the temporal dynamics of card interactions and reduces "Timing Errors" (see Section~\ref{sec:timing-error}).
    \item \textbf{Synergy Analysis:} The core of the response, where the LLM analyzes the synergy between the two cards based on their specific combination. It is important that this analysis distinguishes between the general effects of each card and those that arise only from the specific synergy. This section helps the model focus on the unique interaction of the cards, rather than merely restating their individual effects.
    \item \textbf{Conclusion:} In this step, the LLM summarizes the synergy analysis and categorizes the combined effect as either positive, negative, or neutral. If multiple synergies are identified, the LLM must decide which ones are most significant and provide a final classification. This step resolves any conflicting analyses and clarifies the outcome.
    \item \textbf{Final Score:} The final step involves the LLM providing a single numeric score (1, -1, or 0) that represents the overall synergy between the two cards. This score serves as the final output of the prompt and is used to evaluate the model's performance. However, the entire response, including reasoning, is documented for further error analysis (discussed in Section~\ref{sec:error-classification}).
\end{itemize}

\section{Results}
\label{sec:model-comparison}

In this section, we present the performance of various LLMs on the synergy detection task. As outlined previously, our dataset is highly imbalanced, so we use the F1 score as our primary evaluation metric. We report precision, recall, and F1 scores for each class and on average, as well as the overall accuracy. The F1 score we report is the macro average, which is calculated as the mean of the F1 scores for the three classes (positive, negative, and neutral). All results are summarized in Table~\ref{tab:results}.

\begin{table*}[!t]
\caption{Model comparison\label{tab:results}}
\centering
\renewcommand{\arraystretch}{1.5}
\begin{tabular}{|c||c|c|c|c||c||c|}
\hline
\multirow{2}{*}{Model} & \multirow{2}{*}{Class} & \multirow{2}{*}{Precision} & \multirow{2}{*}{Recall} & \multirow{2}{*}{F1 Score} & \multirow{2}{*}{Accuracy} & \multirow{2}{*}{\shortstack{Invalid \\Response}} \\
& & & & & & \\
\hline
\multirow{4}{*}{\shortstack{GPT-4o-mini}}
& Positive Synergy & 0.34 & 0.69 & 0.46 & \multirow{4}{*}{0.68} & \multirow{4}{*}{0\%}\\
& No Synergy & 0.91 & 0.69 & 0.78 & &\\
& Negative Synergy & 0.12 & 0.23 & 0.16 & &\\
\cline{2-5}
& Average & 0.46 & 0.54 & \textbf{0.47} & &\\
\hline
\multirow{4}{*}{\shortstack{GPT-4o-mini \\ \\  (Fine-tuned)}}
& Positive Synergy & 0.51 & 0.52 & 0.52 & \multirow{4}{*}{0.79} & \multirow{4}{*}{0\%}\\
& No Synergy & 0.88 & 0.87 & 0.88 & &\\
& Negative Synergy & 0.11 & 0.16 & 0.13 & &\\
\cline{2-5}
& Average & 0.50 & 0.52 & \textbf{0.51} & &\\
\hline
\multirow{4}{*}{GPT-4o}
& Positive Synergy & 0.59 & 0.71 & 0.65 & \multirow{4}{*}{0.81} & \multirow{4}{*}{0\%}\\
& No Synergy & 0.92 & 0.84 & 0.88 & &\\
& Negative Synergy & 0.11 & 0.31 & 0.16 & &\\
\cline{2-5}
& Average & 0.54 & 0.62 & \textbf{0.56} & &\\
\hline
\multirow{4}{*}{\shortstack{Gemini 1.0 Pro}}
& Positive Synergy & 0.23 & 0.83 & 0.36 & \multirow{4}{*}{0.42} & \multirow{4}{*}{2.2\%}\\
& No Synergy & 0.91 & 0.34 & 0.50 & &\\
& Negative Synergy & 0.07 & 0.31 & 0.12 & &\\
\cline{2-5}
& Average & 0.41 & 0.49 & \textbf{0.33} & &\\
\hline
\multirow{4}{*}{\shortstack{Gemini 1.5 Flash}}
& Positive Synergy & 0.45 & 0.58 & 0.51 & \multirow{4}{*}{0.76} & \multirow{4}{*}{15.0\%}\\
& No Synergy & 0.89 & 0.80 & 0.85 & &\\
& Negative Synergy & 0.12 & 0.30 & 0.17 & &\\
\cline{2-5}
& Average & 0.49 & 0.56 & \textbf{0.51} & &\\
\hline
\end{tabular}
\end{table*}

One of the models we evaluated is a fine-tuned version of GPT-4O-mini. This model was fine-tuned on 160 examples of synergy questions and responses, following the format described in Section~\ref{sec:prompt-structure-answer}. The examples are carefully selected from cases where our initial results showed errors, while ensuring the inclusion of negative synergies. This fine-tuning aimed to improve performance by addressing common pitfalls observed in the initial model outputs.

For each experiment, we observed that the models, including the fine-tuned version, faced significant challenges in detecting synergies, especially negative synergies. The F1 scores for negative synergies remained consistently low, with values of 0.17 or below across all models tested. This difficulty can be attributed to the inherent complexity of negative synergy examples, which often involve subtle interactions between cards. In many cases, the cards involved do not feature explicit keywords or clear descriptions that easily signal the nature of their interaction, making it more difficult for the model to recognize the synergy.

Additionally, we encountered cases where the models did not follow the expected response format. In these instances, the automatic extraction of the synergy score from the model's output would fail, in which case we resend the prompt to the model up to three times. If the model fails to provide a valid output after three attempts, the response is considered invalid and excluded from the final evaluation. Table~\ref{tab:results} also includes the percentage of excluded cases due to invalid response.

Overall, while the models showed some proficiency in detecting positive synergies, they perform poorly on detecting when there is a synergy, especially a negative synergy. A deeper analysis of the error cases is provided in Section~\ref{sec:error-classification}.

\section{Error Classification}
\label{sec:error-classification}

Upon closer look at the responses from experiments with our best-performing GPT-4o model, we found similar types of errors in the responses. In this section, we classify these errors into different types.

\subsection{Error Patterns}
\label{sec:error-patterns}

Figure~\ref{fig:heatmap} visualizes the errors based on the index of first card played (y axis) and second card played (x axis). We observe two types of error pattern in this visualization:

\begin{figure}
    \centering
    \includegraphics[width=\linewidth]{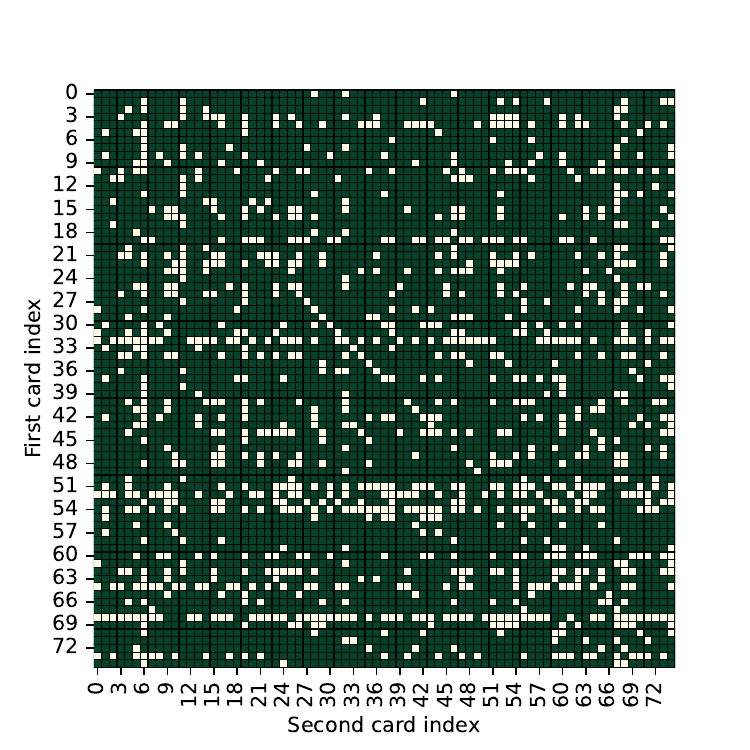}
    \caption{Error patterns based on our experiments using the GPT-4o model. Dark and light colors show successful and unsuccessful classifications.}
    \label{fig:heatmap}
\end{figure}

\subsubsection{Row/Column Patterns}
These patterns show struggles in classifying synergies when specific cards are played first (row pattern) or second (column pattern). This happens most often when they produce an effect that has a significant positive or negative effect, but does not interact with other cards to create a synergy. One such example is the card ``Fiend Fire'', which removes every other card in the player's hand from the deck for a large burst of damage. This large-scale effect causes the model to classify as negative the neutral or no synergy cases.

\subsubsection{Diagonal pattern}

Additionally, we see a a pattern of errors along the diagonal of Figure~\ref{fig:heatmap} where a card is played with a copy of itself. This occurs when playing a card twice causes the amplitude of the given effect to become larger, which the model often misclassifies as a positive synergy.

\begin{figure}
    \centering
    \includegraphics[width=0.7\linewidth]{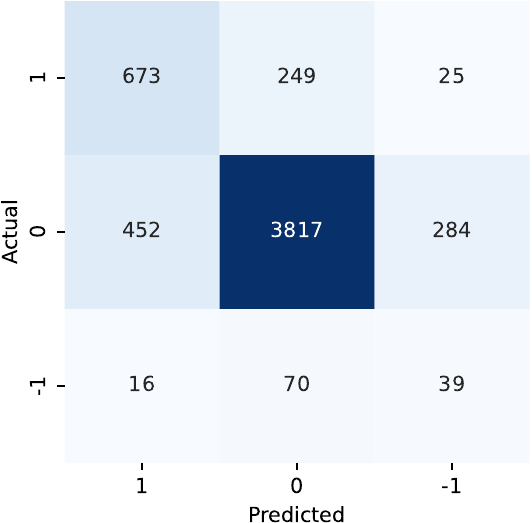}
    \caption{Confusion matrix and error rates for each class.}
    \label{fig:conf-matrix}
\end{figure}

As we previously discussed, negative synergies are often misclassified. Figure~\ref{fig:conf-matrix} visualizes a confusion matrix for the GPT-4o model, where negative synergies are more often classified as no synergy, and sometimes even positive synergy.

Almost all observed error patterns remained relatively consistent as we refined our prompt. Additionally, our experiments using both GPT-4o-mini and GPT-4o showed the same general error patterns.

\subsection{Description Error}
\label{sec:description-error}

In our analysis of the responses from our initial experiments and especially when focusing on the failure cases, we found that a common type of error is when the description of the card or the order of its effects is slightly altered in the response.

To minimize the occurrence of this type of error, we require the LLM to include a word-for-word repetition of the card descriptions in its response as the first step to the chain-of-thought reasoning. Even small changes in wording can significantly alter the meaning and effect of a card, so this recitation must be exact.

We found that the few-shot examples can significantly help in this case if they also follow a word-for-word copy of the description. Even a slight, insignificant change in reciting the card descriptions in the few-shot examples can significantly increase the frequency of this error.

\subsection{Timing Error}
\label{sec:timing-error}

The Timing Error is made when there is a failure to understand the correct order of events. There are four main forms of Timing errors: Timing, Interruption, Continuity, and Immediacy.

\subsubsection{Timing Error}
The first error of Timing is when the order of effects of two or more cards is mixed up. For example, if a card that does 5 damage is played, and another card that increases future damage done is played afterward, a timing error may occur if the LLM states that the 5 damage gets amplified by the second card. This most often occurs when a synergy is presented out of order and would become more likely with more complex cards.

\subsubsection{Interruption Error}
Interruption occurs when the effect of one card is interrupted by the play of another card before the first card's effect finishes. This error commonly occurs when the play of one card prevents the play of the second card for a significant amount of time, so the LLM tries to inject the second card early.

% possibly including this in the appendix? If we want to include it here it should also have card descriptions
% An example of this happens with the Fiend Fire card mentioned above. If played first, the LLM would sometimes attempt to insert the second card played into the middle of Fiend Fire's effect so the second card would not be removed. This would often cause a false positive.

\subsubsection{Continuity Error}
Continuity errors are rare cases where the LLM simply neglects to account for the implications of a previously played effect. This could take the form of playing a card directly after stating that it would be impossible to play or stating that an effect is in play but never mentioning it again.

\subsubsection{Immediacy Error}
Immediacy errors occur when the LLM either overvalues or undervalues how soon an effect, synergistic or not, occurs. In our GPT-4o experiments, this most often took the form of a negative synergy prediction when the two cards could not immediately produce a synergy or be played. For example, the card ``Demon From'' adds a flat amount of damage to all attack cards every turn. This effect is positively synergistic with almost all attack cards, but due to the need to wait before playing the attack card, the response often predicts a negative synergy.

% What did we change about the promp? Also I think if we're including this it should have more explanation
% In order to adress the litany of timing errors, we tweaked the prompt and requested the LLM consider more or less immediate synergies, yet that only caused the problems to shift between positive and negative bias.

\subsection{Game State Error}
\label{sec:game-state-error}
Game State errors are cases when the LLM creates an artificially optimal or suboptimal game state, and considers the synergistic effect of one of the cards in that state as a synergy between the two cards. This type of error directly relates to the type of challenge discussed in Section~\ref{sec:challenge-external-factor}. Game State errors most often occur as a result of the first card having extensive requirements or the combination of the two cards requiring some form of setup.
% possible appendix
% A good example of this is with the card Dropkick is played first. Dropkick has a unique effect when the enemy has the Vulnerable status. While we would like the LLM to take the unique effect into account when determining synergy, in this case, the LLM would assume that the enemy has the Vulnerable status, causing Dropkick to become synergistic when played first with any damaging card, as Vulnerable increases damage by 50\%. It is important to note that game state errors tend to also be the source of most false successes where the LLM returns the correct value, but has faulty reasoning. To address this, we reviewed our dataset to make sure we were being reasonable in all of our initial game states and informed the LLM to focus only on the synergy between the two given cards, but it continued to make Game State Errors when cards explicitly required certain states to function optimally.

\subsection{Rule Error}
\label{sec:rule-error}
Rule errors are cases where the game rules are not applied correctly by the LLM to its hypothetical scenario. This type of error is more common with ambiguous or unique effects (also relates to Section~\ref{sec:challenges-rule-interpretation}). This type of error would often cause a card to have a high error rate as all interactions with that card were likely to fall victim to the same error type, creating some of the row and column patterns in Figure~\ref{fig:heatmap}.

Another common Rule error is seen when a card that has been played is believed to still be in the player's hand, causing new restrictions and benefits that would not occur in the game. To address these errors, we included clear definition of some of these rules and details on certain unique effects in the prompt. Despite these changes, misinterpretation and new points of contention arose, causing the Rule errors to remain common.

\subsection{Skip Error}
\label{sec:skip-error}

This type of error occurs when some steps from the chain-of-thought reasoning are skipped. For example, the response may exclude the "Order of Events" section entirely, which can result in Timing errors. We found this type of error can occur even if the prompt clearly states what exact sections should be included in the response. However, using specific section names and including these names with clear distinctions in the few-shot examples resulted in the prevention of this error.

\subsection{Similarity Error}
\label{sec:similarity-error}
Similarity errors occur when two cards have similar effects, which may lead the LLM to mistakenly confuse this situation with cases where one card's effect is amplified due to a synergy. This often occurs when one of the two cards has a much larger effect than the other, a unique effect that is distinctly positive or negative, or when the same card is played twice. The result of such errors is both the diagonal pattern and the more sporadic errors that pepper Figure~\ref{fig:heatmap} as opposed to being specific to one card. For instance, when the cards both simply deal damage to the enemy, no synergy should be detected. However, the LLM's response states that since the second card made more of the first cards effect occur, there must be a synergy. To reduce its frequency, we specifically mention in the prompt to consider whether or not the synergy effect in question still persists if one of the cards is removed from the scenario. If so, the effect is not synergistic and is only a positive or negative effect for one of the cards. However, this type of error persisted despite these changes.

\section{Findings and Future Work}

In the previous sections, we demonstrated that LLMs perform poorly on our synergy detection dataset, suggesting that they may struggle with understanding complex game rules. We also categorized the types of errors commonly made by LLMs when performing this task. In this section, we discuss several potential avenues for future work aimed at improving LLM performance in synergy detection.

\subsection{Prompt Design Improvements}

Our experiments revealed that prompt design plays a critical role in improving LLM performance. Specifically, we identified a few strategies that could reduce common errors. First, to prevent Description errors, we recommend explicitly including a step in the prompt where the LLM restates the description of the cards involved in the synergy. This step ensures that the LLM accurately captures the intended meaning of each card, minimizing the risk of misinterpretation. Additionally, we found that including a word-for-word copy of the cards involved in the synergy in the few-shot examples helps prevent the LLM from altering the description.

Second, to address Timing errors, we suggest incorporating a dedicated step in the chain-of-thought reasoning to reinforce the exact order of events. Ensuring that the LLM reiterates the sequence of actions can improve its ability to determine the correct temporal relationships between synergies.

Finally, to reduce Skip errors, we propose assigning a unique name to each step in the chain-of-thought process. Including these names in the few-shot examples can help guide the LLM through each reasoning stage, reducing the likelihood of omitting critical steps in the response.

We did not see a significant improvement when using prompts to prevent other error types, for example by highlighting important rules regarding card draw and hand logic in the prompt.

Notably, we experimented with changing the keyword that refers to the synergy effect. Interestingly, for the GPT-4o-mini model, our initial experiments shows that the word ``Synergy'' carries a bias that negatively impacted the results. We tested other keywords, including both meaningful and meaningless alternatives, but found no significant difference in their effects. Based on these findings, we opted to use the word "Combo" to refer to the synergy effect in our experiments. The complete prompt and further details can be found in our previously linked GitHub repository.

\subsection{Fine-Tuning}

We observed that fine-tuning the LLM on Question-Answer examples provided some improvements, particularly in preventing Description and Skip errors. However, the overall effect was modest, and the model continued to struggle with synergy detection even with 160 fine-tuning examples. One possible reason for this is that small variations in phrasing can lead to significant changes in meaning, making it difficult to generalize across scenarios.

Nonetheless, fine-tuning have shown a great effect on preventing Description and Skip errors. As expected, fine-tuning also improves the ability of the LLM to follow the correct response format way more effectively compared to describing the format in the prompt or the few-shot examples, preventing invalid responses.

\subsection{Temporal Reasoning}

Our results also indicate that LLMs struggle with predicting the order of effects in synergy detection tasks, pointing to a significant gap in their temporal reasoning abilities. This suggests that future work may explore temporal benchmarks \cite{temporal-benchmark, temporal-improvement} to focus on improving LLMs' understanding of temporal relationships in complex tasks. Tan et al. suggest that fine-tuning on temporal reasoning datasets prior to training on the main task dataset can improve performance \cite{temporal-improvement}. This could be a promising direction for enhancing LLMs' ability to handle tasks that require temporal reasoning, such as synergy detection.

\subsection{Handling Complex Errors}

We encountered several challenging errors that require further investigation. Specifically, Game State errors, Rule errors, and Amplification errors remain significant hurdles for accurate synergy detection. One potential strategy for addressing these challenges is the development of a self-refinement process, where the LLM is tasked with evaluating and providing feedback on its own responses. This could allow the model to iteratively improve its performance, but it may come at the cost of higher computational resources. Exploring this approach could lead to substantial improvements, but further research is needed to determine its feasibility and effectiveness.

\subsection{Alternative Output Formats}

While improvements in prompt design, fine-tuning, and model architecture can reduce Skip errors, there is still room for exploration in terms of output formatting and preventing invalid responses, especially for the Gemini models. Although errors of this type are less common with the use of few-shot examples and stronger models, further progress could be made by experimenting with alternative text-based formats. For example, using a structured format like JSON for LLM outputs may help enforce consistency and clarity in the model’s reasoning process. This avenue could provide a valuable alternative to pure text-based outputs and warrants further investigation.

\section{Conclusion}

In this paper, we explored the performance of large language models (LLMs) in reasoning about complex rule interactions in the context of card games, using \textit{Slay the Spire} as a case study. Our analysis revealed that while LLMs perform well in identifying non-synergistic card pairs, they face significant challenges when it comes to detecting positive and negative synergies. These results underscore the complexity of reasoning within rule-based environments and highlight the gaps in LLM capabilities when faced with rule interactions. The errors observed in synergy detection suggest that further refinement is needed to enable models to more reliably reason about the consequences of rule interactions.

Moreover, our findings also suggest that the task of synergy detection could be an important benchmark for future research on LLMs, particularly in the areas of temporal reasoning, rule-based decision-making, and fine-tuning. This work opens the door for future studies that could explore methods to better equip LLMs for understanding and interacting with complex, dynamic systems—both in the context of games and beyond.

{\appendices
\section{LLM Prompt}

This section outlines the prompt used in our experiments. The conversation consists of a series of user and assistant (LLM) messages. It is important to note that the assistant's responses in this section (marked as Injected) are not generated, but are injected into the conversation similar to few-shot examples.

The conversation begins with the Context Block, which includes the game description, relevant keywords, and the definition of synergy. Following this, the final message in the conversation marks the start of the few-shot examples. Eight examples are then inserted into the conversation before the final prompt, which serves as the main query.

\begin{convbox-prompt}
You are studying a new card game. In this game, the player has a deck of cards.\\
The game is single player and is played in a number of turns. The game is played against one or more enemies.\\
The player has some amount of HP. If the player's HP reaches zero, the player loses.\\
Each of the enemies also has some amount of HP. Any enemy with zero HP dies and does not affect the game anymore. If all the enemies die, the player wins the game.\\
At the start of every turn, each enemy also shows an intention. This intention describes what action that enemy performs after the player ends their turn. Example enemy actions are "deal 5 damage" or "gain 5 Block".\\
The game has four card piles: draw pile, discard pile, exhaust pile, and your hand. Initially, the player starts the game by shuffling the deck of cards into the draw pile. The other piles are empty.\\
Every turn, these actions are performed in this exact order:\\
1. The player gains some amount of energy. This amount is usually 3, but some items or cards in the game can change this amount.\\
2. The player draws some number of cards from the draw pile into their hand. This number is usually 5, but some items or cards in the game can change this amount.\\
* If there are not enough cards in the draw pile, all the cards in the discard pile shuffle back into the draw pile. Then, the remaining cards are drawn.\\
3. The player plays as many cards as they want, as long as they have enough energy to play the cards. Then, they end their turn.\\
4. When the player ends their turn, all the remaining cards in their hand are moved to the discard pile. New cards will be drawn next turn.\\
5. When the player ends their turn, the enemies perform the actions shown as their intention at the start of the turn.\\
Any questions so far?
\end{convbox-prompt}

\begin{convbox-response-inj}
No questions so far! The rules of the card game are clear. The mechanics regarding the turn structure, energy management, card drawing, and enemy actions are all well understood.
\end{convbox-response-inj}

\begin{convbox-prompt}
OK, I want to explain some terms from the game before asking questions.\\
- Card Type: Attack, Skill, and Power are the three Card Types. Power cards are played and then removed from that battle, but Attack or Skill cards move to the discard pile if they are played. We also have Status or Curse cards, which are unplayable and, most of the time, are the cards that disadvantage the player in some way.\\
- Block: Prevents a certain amount of damage until the next turn. For example, if the player gains 5 Block, and the enemy deals 7 damage to them that turn, the player loses only 2 HP and all their Block. It only lasts one turn, so if there is any remaining Block on the enemy or the player, the Block is removed right before they perform their next set of actions. Additionally, if a card takes away the player’s HP directly, the Block does not prevent the HP loss.\\
- Upgrade: When a card is upgraded, its effects are enhanced and it sometimes costs less energy.\\
- Vulnerable: The affected target takes 50\% more damage from Attack damage. Note that this only applies to damage from an Attack type card or an enemy Attack action. The number of Vulnerable shows how many turns it will last.\\
- Weak: The affected target deals 25\% less damage with Attacks. Note that this only applies to damage coming from an Attack type card or an enemy Attack action. The number of Weak shows how many turns it will last.\\
- Strength: Adds 1 additional damage to Attacks from the affected entity per stack of Strength. Note that this only applies to damage coming from an Attack type card or an enemy Attack action. This only applies to the Attack damages comming from the entity that gained the Strength. If a target Loses Strength while they have 0 Strength, they strength becomes negative, making them deal 1 less damage per Stack. This also only applies to the Attack damages comming from the entity that gained the Strength.\\
- Exhaust: Remove the affected card until the end of combat. You do not automatically draw a card to replace it.\\
- Wound: An unplayable card with no effects.\\
- Ethereal: If an Ethereal card is in the player's hand at the end of the turn, it is Exhausted.\\
- Dazed: An unplayable card that is also Ethereal.\\
- Innate: Start each combat with this card in your hand.\\
- Burn: An unplayable card. If the card is in the player's hand at the end of their turn, the player will take 2 damage.\\
- X cost card: If a card's cost is X, it means that when the card is played, it will consume all the available energy the player has. This number of energy consumed is X, and the effect of the card is often described in terms of X.\\
- When a card says "At the start of your turn ..." or "At the end of your turn ...", it means this effect will happen every single turn at the start of the player's turn or at the end of the player's turn.\\
- When you play a card, that card is no longer in your hand and will not be affected by "in your hand" effects. It also will not contribute to your current hand size after being played.\\
- When you *gain* an effect, like Vulnerable or Strength, the effect is applied to the player. When you *apply* an effect, the effect is applied to an enemy.\\
- Effects are not applied retroactively. So, for example, if I gain Strength *after* dealing Attack damage, the Attack damage does not get amplified by the Strength effect. Additionally, if you play a Power card, it will not impact any cards before it was played but does apply its effect to any cards afterward.\\
- When a random target is picked, the random target changes each time the effect happens.\\
Any questions so far?
\end{convbox-prompt}

\begin{convbox-response-inj}
No questions at the moment. I'm ready to give feedback on specific card designs or mechanics you have in mind!
\end{convbox-response-inj}

\begin{convbox-prompt}
Now, I want you to help me understand card combo effects.\\
A combo effect is when two cards have some additional effect when played together, as opposed to playing them separately.\\
Let's say we have two cards, card A and card B. Each card has positive and negative effects and can be strong or weak in different situations.
Suppose we play card A in some scenario with other cards. Suppose we play card B in a scenario with some other cards.\\
Is there a scenario S in which playing card A, then card B, would give us additional advantages or disadvantages compared to the effects of cards A and B separately?\\
If, in most scenarios, there are some extra advantageous effects of playing the cards together, we call this a positive combo. A positive combo can also occur when the effects from card A prevent some of the negative effects of card B or vice versa.\\
If, in most scenarios, there are some additional disadvantageous effects when A and then B are played (compared to playing them separately), we call this a negative combo. A negative combo can also occur when the effects from card A prevent some of the positive effects of card B or vice versa.\\
Note that:\\
- When we say "play card A, then card B", this can mean playing B immediately after playing A, or after playing extra cards in between, or even after passing a few turns. However, the effects of the extra cards played should not be considered for the combo.\\
- When we talk about combo between card A and card B, ONLY CONSIDER combo BETWEEN A AND B. For example, if card A has a lot of combo with other cards, it doesn't matter because playing card A with other cards would still give us those effects. We want to see what additional effects we get from A and B TOGETHER.\\
Now, given two cards, I want you to first describe their combo, if any. I want you to think about this effect very carefully. Again, some cards might not have a combo at all. When trying to understand the combo, we want to think about the effect of the cards played in the exact given order, immediately or after waiting some number of turns or playing some extra cards. Let's also assume that you have enough energy to play them in the same turn if needed, but if you think there are some energy considerations from one card to another that can be seen as a combo, I want you to describe it thoroughly.\\
Second, given this description, I want you to tell me if the combo is positive or negative or if there is no combo.\\
Finally, I want you to give me a final number at the end of your response in a single line.\\
0: no combo\\
1: positive combo\\
-1: negative combo\\
When I give you the card descriptions, I want these exact three things. Make sure to describe each part and end your response by writing the final number in a single line.\\
Any questions?
\end{convbox-prompt}

\begin{convbox-response-inj}
I’m ready to analyze the combos for you. Please provide the descriptions of card A and card B.
\end{convbox-response-inj}

\begin{convbox-prompt}
Ok, let's think! Pay attention to the exact description of each card and the correct order of differnt effects when playing it.\\
When evaluating combo between two cards, note that the scenario that you imagine should rely on these two cards being played in this exact order.\\
In other words, if you remove one card from the scenario, the combo that your describing shouldn't exist anymore.\\
Let's say we have:\\
Card 1 (Attack Type) - Cost 4: "Costs 1 less energy for each time you lose HP in combat. Deal 18 damage."\\
Card 2 (Attack Type) - Cost 2: "Deal 8 damage. Apply 2 Vulnerable."\\
What is the combo effect of playing Card 1, then Card 2?
\end{convbox-prompt}

\section{Error Examples}

This section presents several examples to illustrate common errors discussed in the paper. The examples are taken directly from the responses of the GPT-4o model. In each case, the user prompt provided is the first message that follows the few-shot examples.

The examples may use the following terms:
\begin{itemize}
    \item Exhaust: Remove the affected card from the deck until the end of combat.
    \item Vulnerable: The affected target takes 50\% more damage from Attack damage. Note that this only applies to damage from an Attack type card or an enemy Attack action. The number of Vulnerable shows how many turns it will last.
    \item Weak: The affected target deals 25\% less damage with Attacks. Note that this only applies to damage coming from an Attack type card or an enemy Attack action. The number of Weak shows how many turns it will last.
\end{itemize}

\subsection{Timing Error}

The following example is a case of immediacy error, where an existing synergy is correctly identified but ultimately disregarded in the final synergy score due to its long-term nature. This is shown in the last few sentences of the conclusion section of the response.

\begin{convbox-prompt}
Nice job! Let's say we have:\\
Card 5 (Power Type) - Cost 0: At the start of your turn, lose 1 HP and draw 1 card.\\
Card 6 (Attack Type) - Cost 2: Exhaust all cards in your hand. Deal 7 damage for each Exhausted card. Exhaust.\\
What is the combo effect of playing Card 5, then Card 6?
\end{convbox-prompt}

\begin{convbox-response}
\#\#\# Card Descriptions:\\
- **Card 5:**\\
- Type: Power\\
- Cost: 0\\
- "At the start of your turn, lose 1 HP and draw 1 card."\\

- **Card 6:**\\
- Type: Attack\\
- Cost: 2\\
- "Exhaust all cards in your hand. Deal 7 damage for each Exhausted card. Exhaust."\\

\#\#\# Order of Events\\
- **Playing Card 5 First:**\\
  - Card 5 is a Power-type card that sets an ongoing effect: at the start of each turn, you lose 1 HP and draw an extra card.\\
  
- **Playing Card 6 Next:**\\
  - Card 6 exhausts all cards currently in your hand and deals 7 damage per exhausted card.\\

\#\#\# Analyzing the Combo:\\
- Playing Card 5 sets up an effect where you consistently draw an additional card each turn, albeit at the cost of 1 HP.\\
- This increases your hand size on subsequent turns, which could enhance Card 6's damage output if it is played with a larger hand size due to the extra drawn card.\\
- The HP loss from Card 5 does not directly affect Card 6 or its mechanics.\\

\#\#\# Conclusion:\\
In the turn you initially play these cards, Card 5 doesn't directly impact the immediate use of Card 6. However, its ongoing effect can increase the potential damage of Card 6 on a future turn by allowing an additional card to be in your hand, therefore potentially increasing the total damage dealt by Card 6. This sequence reveals a potential positive synergy over the long-term gameplay instead of immediate synergy, but there is no direct combo effect when considering Card 5 is played right before Card 6 within a single turn.\\

Final score:\\
0
\end{convbox-response}

\subsection{Rule Error}

In the following example, the LLM makes a mistake by failing to adhere to a fundamental rule of the game. After playing Card 5, the model incorrectly assumes that Card 5 remains in the player's hand and can be targeted by Card 6, despite the response clearly stating that the cards "were played in the same turn." This oversight ignores the fact that playing a power card removes it from the game, a rule that is explicitly stated in the game's instructions.

\begin{convbox-prompt}
Nice job! Let's say we have:\\
Card 5 (Power Type) - Cost 1: Whenever you lose HP from a card, gain 1 Strength.\\
Card 6 (Skill Type) - Cost 1: Create a copy of an Attack or Power card in your hand.\\
What is the combo effect of playing Card 5, then Card 6?
\end{convbox-prompt}

\begin{convbox-response}
\#\#\# Card Descriptions:\\
- **Card 5:**\\
  - Type: Power\\
  - Cost: 1\\
  - Effect: Whenever you lose HP from a card, gain 1 Strength.\\

- **Card 6:**\\
  - Type: Skill\\
  - Cost: 1\\
  - Effect: Create a copy of an Attack or Power card in your hand.\\

\#\#\# Order of Events\\
- **Playing Card 5 First:**\\
  - Activates a persistent effect where you gain 1 Strength whenever you lose HP from a card.\\
  - Establishes an ongoing benefit for the duration of the combat.\\

- **Effect on Card 6:**\\
  - Allows you to create a copy of an Attack or Power card currently in your hand.\\
  - Assuming Card 5 is still in your hand (if it were played in the same turn or sequence), you can create a duplicate of Card 5.

\#\#\# Analyzing the Combo:\\
- The key interaction here is that playing Card 6 after Card 5 provides an option to create another copy of Card 5, potentially allowing for double or more instances of its effect.\\
- This means that for future turns in combat, every instance of taking HP loss due to a card effect will grant 2 or more Strength (depending on how many copies you create and use).\\
- This creates a synergistic interaction, as the Power card’s effect can stack for multiplied benefits, enhancing your attack power significantly over time.

\#\#\# Conclusion:\\
The sequence of playing Card 5 followed by Card 6 leads to a positive combo effect due to the potential for stacking the Strength-gaining effect, significantly boosting your damage capabilities.\\

Final score:\\
1
\end{convbox-response}

\subsection{Similarity Error}

Consider the card in the scenario below, which applies a Weak effect to an enemy. Playing two copies of this card does not create any synergy, as the effects simply stack additively. However, because the second instance of playing the card applies the same type of effect as the first, the LLM may confuse the amplification of the effect with synergy. In reality, the effects are added linearly, and there are no synergistic interactions between the two instances. This type of error can occur both when playing the same card twice or when playing two cards with similar effect types.

The example also highlights a Rule Error, as the LLM states that applying two stacks of Weak results in the target dealing 50\% less damage. However, the game rules specify that stacking Weak increases its duration, not its damage penalty.

\begin{convbox-prompt}
Nice job! Let's say Card 5 is defined as:\\
Card 5 (Skill Type) - Cost 0: Apply 1 Weak to ALL enemies. Exhaust.\\
Let's say we have at least two of this card in our deck.\\
What is the combo effect of playing Card 5, then another one of Card 5 again?
\end{convbox-prompt}

\begin{convbox-response}
\#\#\# Card Descriptions:\\
- **Card 5:**\\
- Type: Skill\\
- Cost: 0\\
- Apply 1 Weak to ALL enemies. Exhaust.\\

\#\#\# Order of Events\\
- **Playing Card 5 First:**\\
- Applies 1 Weak to all enemies, reducing their Attack damage by 25\% for one turn.\\

- **Playing Another Card 5 Next:**\\
- Applies an additional 1 Weak to all enemies, stacking with the first, resulting in 2 Weak on each enemy.\\

\#\#\# Analyzing the Combo:\\
- The effect of Weak stacks, meaning enemies now deal 50\% less damage for the duration of the effect, which is more beneficial than playing a single instance.\\
- This sequence is effective at compounding the advantage of the Weak effect on all enemies, magnifying its impact by layering the debuff.\\

\#\#\# Conclusion:\\
The sequence effectively creates a positive combo due to the cumulative application of Weak, enhancing the player's defensive position by significantly reducing enemy attack damage over the turns the Weak effects last.\\

Final score:\\
1
\end{convbox-response}
}

\bibliographystyle{IEEEtran}
\bibliography{main}

\newpage

\vfill

\end{document}